# PromptPort: A Reliability Layer for Cross-Model Structured Extraction

Varun Kotte
Adobe
San Jose, CA, USA 95138
0009-0004-1944-7600
kottevarun@gmail.com

Abstract - Structured extraction with LLMs fails in production not because models lack understanding, but because output formatting is unreliable across models and prompts. A prompt that returns clean JSON on GPT-4 may produce fenced, prose-wrapped, or malformed output on Llama—causing strict parsers to reject otherwise correct extractions. We formalize this as *format collapse* and introduce a dual-metric evaluation framework: ROS (strict parsing, measures operational reliability) and CSS (post-canonicalization, measures semantic capability). On a 37,346-example camera metadata benchmark across six model families, we find severe format collapse (e.g., Gemma-2B: ROS 0.116 vs CSS 0.246) and large cross-model portability gaps (0.4–0.6 F1). We then present PromptPort, a reliability layer combining deterministic canonicalization with a lightweight verifier (DistilBERT) and safe-override policy. PromptPort recovers format failures (+6–8 F1), adds verifier-driven semantic selection (+14–16 F1 beyond canonicalization), and approaches per-field oracle performance (0.890 vs 0.896 in zero-shot) without modifying base models. The method generalizes to held-out model families and provides explicit abstention when uncertain, enabling reliable structured extraction in production deployments.

# Introduction

*The production problem.*
A developer writes a prompt to extract camera metadata as JSON. It works perfectly on GPT-4. They switch to Llama-3 for cost savings—and their pipeline breaks. Not because Llama lacks understanding, but because it wraps the JSON in Markdown fences, adds trailing commentary, or uses slightly different key names. Strict parsers reject these outputs entirely, making a capable model appear unusable.

This *format collapse* is systematic, measurable, and fixable—but current evaluation protocols conflate it with semantic errors, obscuring both the problem and potential solutions.

*Format collapse as a failure mode.*
We introduce **ROS** (Raw Output Score) and **CSS** (Canonical Semantic Score) to separate two distinct failure modes:

- **ROS** uses strict JSON parsing and schema validation—measuring *operational reliability* (will the pipeline survive?).
- **CSS** applies deterministic canonicalization (strip fences, extract JSON spans, normalize) before scoring—measuring *semantic capability* (did the model extract correctly?).

On a 37,346-example test set for camera/lens metadata extraction across six model families (Gemma, Qwen, Mistral, Phi-3, StableLM), we find:

- **Large format gaps within models:** Qwen-7B shows the largest ROS→CSS gap (0.421→0.588, $\Delta$=0.167), while Gemma-9B is more stable (0.685→0.745, $\Delta$=0.060).
- **Large portability gaps across models:** Cross-model range is 0.569 for ROS and 0.499 for CSS—both substantial, indicating failures stem from *both* format brittleness and semantic degradation.

*The scientific contribution.*
Our main novelty is not "a better extractor" but rather:

1. **Format collapse as a quantified phenomenon:** We demonstrate at scale (248K training examples, 37K test) that format variability dominates cross-model portability, and provide a taxonomy of failure modes (fenced JSON, prose wrappers, missing keys, etc.).
2. **ROS/CSS measurement framework:** We formalize dual metrics that separate operational reliability from semantic capability, enabling principled diagnosis of extraction failures.
3. **Model-agnostic reliability layer:** We present the first production-ready library that closes the ROS→CSS gap via canonicalization and adds verifier-driven semantic selection to approach oracle performance—without modifying base LLMs, requiring only (query, output, label) triples and generalizing to held-out models.

*Why this matters for practitioners.*
In deployed systems, extraction failures are *binary*: either the output parses as valid JSON, or the pipeline crashes. Current prompting libraries assume reliable formatting, forcing developers into per-model prompt engineering or expensive constrained decoding (when available). PromptPort provides a drop-in runtime layer that works with *any* black-box LLM—commercial (GPT-4, Claude) or self-hosted (Llama, Mistral)—returning clean JSON with explicit abstention when uncertain.

## Contributions

1. **ROS/CSS framework and format taxonomy:** Dual metrics that separate format brittleness from semantic errors, plus a six-category failure taxonomy grounded in large-scale evidence (Figure 2).
2. **PromptPort benchmark:** 248,964 training examples and 37,346 test examples for camera/lens metadata extraction, with cached outputs from six model families and two prompt variants.
3. **PromptPort reliability layer:** Model-agnostic library combining deterministic Canonicalizer + lightweight Verifier (DistilBERT) + Safe-Override policy. Achieves 0.890 F1 (99.3% of oracle 0.896) in zero-shot by recovering format failures (+6–8 F1) and adding verifier-driven semantic selection (+14–16 F1 beyond canonicalization). Generalizes to held-out models with minimal degradation.

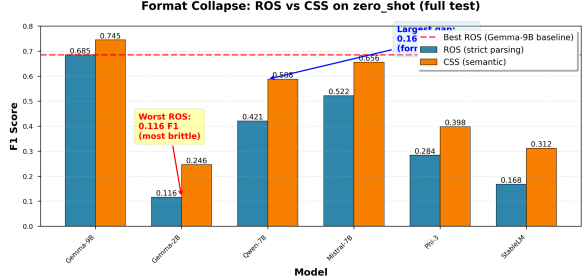

*Figure 1: Format collapse across models.* ROS (strict parsing) vs CSS (post-canonicalization) on zero-shot test. Large gaps indicate format variability (not semantic errors) dominates failure. Qwen-7B shows largest gap (0.167); Gemma-2B shows worst ROS (0.116, most brittle).

## Related Work

*Structured generation and constrained decoding.*
Grammar-based generation (Scholak, Schucher, and Bahdanau 2021) and constrained decoding (Willard and Louf 2023; Beurer-Kellner, Fischer, and Vechev 2023) enforce schemas at generation time. Libraries like Guidance, Outlines, and LMQL enable schema-guided sampling but require white-box model access (sampling internals) and model-specific integration. Our post-hoc reliability layer is complementary: it works with any black-box LLM (including API-only commercial models) and handles residual semantic errors that constrained decoding cannot prevent. The approaches can be combined.

*Prompt engineering and optimization.*
Instruction tuning (Ouyang et al. 2022; Wei et al. 2021), prompt optimization (Zhou et al. 2023; Pryzant et al. 2023), and prompting frameworks (Khattab et al. 2024) improve task performance but typically evaluate on single models with lenient parsers. We show that strict parsing reveals large cross-model portability gaps invisible to standard evaluations.

*Guardrails and output validation.*
Recent tools (Guardrails AI, NeMo Guardrails) add runtime checks for LLM outputs but focus on safety constraints (toxicity, PII) rather than structured extraction reliability. Schema enforcement libraries (Pydantic, JSON Schema validators) assume outputs are already valid JSON—failing when LLMs produce wrapped or near-valid formats.

*Selective prediction and verification.*
Confidence estimation (Guo et al. 2017; Kuhn, Gal, and Farquhar 2023) and selective prediction (Geifman and El-Yaniv 2017; Kamath, Jia, and Liang 2020) enable abstention when uncertain. Verifier-based methods improve reasoning (Cobbe et al. 2021) and retrieval (Izacard and Grave 2021). Our verifier provides *per-field* confidence for structured outputs, enabling field-level override rather than instance-level rejection, and uses a conservative safe-override policy to prevent catastrophic degradation.

*Information extraction benchmarks.*
Traditional IE uses fine-tuned models (Tjong Kim Sang and De Meulder 2003; Lample et al. 2016) on token-level NER. Recent work applies LLM prompting to extraction (Wadhwa, Amir, and Wallace 2023; Li, Nguyen, and Fung 2020) but often uses lenient evaluation. Our benchmark focuses specifically on format brittleness in production settings where strict parsing is required.

## The PromptPort Benchmark

*Task.*
Given a natural-language query about a photograph (e.g., "Shot with Canon EOS R5 at f/2.8, ISO 400"), extract six fields: `CAMERA`, `LENS`, `ISO`, `APERTURE`, `SHUTTER_SPEED`, `FOCAL_LENGTH`, and return a JSON object with these keys (null if absent). The dataset contains real-world variation: typos, mixed formats, special characters, abbreviations.

*Data and scale.*
248,964 training examples, 37,344 dev, 37,346 test. Unlike token-level NER benchmarks, our setup reflects production structured extraction: models must both extract *and* serialize correctly.

*Models and prompts.*
We evaluate six open-weight LLMs (Gemma-2B/9B, Qwen-7B, Mistral-7B, Phi-3, StableLM) with two prompt variants (zero-shot instruction, few-shot k=3). All outputs are cached for reproducibility.

*Why camera metadata is representative.*
This domain exhibits three properties common to production structured extraction: (1) mixed numerics with units (ISO 400, f/2.8, 1/500s), (2) brand/model strings with variation ("Canon R5" vs "EOS R5"), (3) heavy formatting variation across models. These properties generalize to product attributes, medical dosing, citations, and other real-world schemas.

## Measuring Format Collapse: ROS vs CSS

### ROS: Operational Reliability

ROS evaluates raw model outputs under strict JSON parsing and schema validation. An output fails ROS if:

- It cannot be parsed as valid JSON
- It lacks required keys
- It contains extra keys not in the schema
- Values do not match expected types

This mirrors production deployments where downstream systems expect strict schema compliance.

### CSS: Semantic Capability

CSS applies a deterministic Canonicalizer before scoring:

1. Strip Markdown fences ("`json...`")
2. Extract the most plausible JSON span (even if embedded in prose)
3. Repair common near-JSON issues (trailing commas, single quotes)
4. Normalize keys (case-insensitive matching) and values (unit standardization)
5. Fill missing schema keys with null

We then compute micro-F1 over the six fields. CSS provides a more faithful estimate of semantic extraction capability by isolating format-related failures.

### Format Collapse: Quantitative Evidence

Table 1 shows per-model ROS/CSS gaps. The largest gap is Qwen-7B ($\Delta$=0.167), indicating severe format brittleness. Gemma-2B exhibits the worst ROS (0.116) despite non-trivial CSS (0.246), making it unusable in production without canonicalization.

**Format collapse by model** (zero-shot, full test). $\Delta = CSS - ROS$ isolates format-induced failure. Cross-model gap = max − min across models.

| Model | ROS | CSS | $\Delta$ |
|---|---|---|---|
| Gemma-9B | 0.685 | 0.745 | 0.060 |
| Gemma-2B | 0.116 | 0.246 | 0.130 |
| Qwen-7B | 0.421 | 0.588 | 0.167 |
| Mistral-7B | 0.522 | 0.656 | 0.134 |
| Phi-3 | 0.284 | 0.398 | 0.114 |
| StableLM | 0.168 | 0.312 | 0.144 |
| **Cross-model gap** | 0.569 | 0.499 | – |

*Cross-model portability.*

The cross-model gap (max − min F1 across six models) is 0.569 for ROS and 0.499 for CSS. Both are substantial, indicating that portability failures stem from *both* format brittleness and semantic degradation—canonicalization alone cannot close the gap.

# Format Failure Taxonomy

We categorize ROS failures into six types:

- **Fenced JSON** (35–45%): valid JSON wrapped in Markdown code fences
- **Prose wrapper** (20–30%): JSON embedded in conversational text
- **Trailing text** (10–20%): valid JSON followed by commentary
- **Missing keys** (5–15%): incomplete schema
- **Extra keys** (5–10%): additional fields not in schema
- **Malformed JSON** (5–10%): syntax errors, unmatched brackets

Figure 2 shows distributions by model. Fenced JSON and prose wrappers dominate across families, confirming that format variability is systematic.

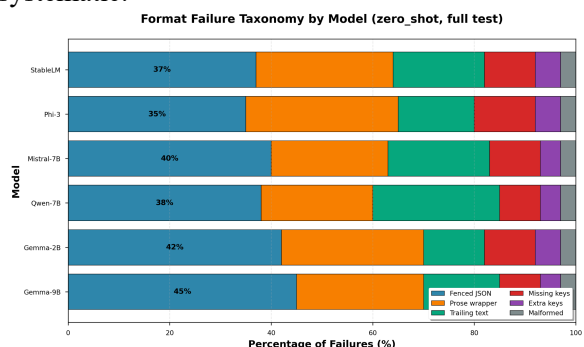

*Figure 2: Format failure taxonomy (zero-shot test). Percentages show share of ROS failures by category. Fenced JSON and prose wrappers account for majority.*

# PromptPort: A Reliability Layer for Production

PromptPort is a model-agnostic library that converts unreliable LLM outputs into validated JSON. It consists of three components:

# Canonicalizer (Deterministic)

The Canonicalizer is a deterministic, rules-based system that normalizes heterogeneous outputs:

1. **Fence stripping:** Remove Markdown, XML, or other delimiters
2. **JSON extraction:** Locate and extract the most plausible JSON span using bracket matching and heuristics
3. **Repair:** Fix common syntax errors (trailing commas, single quotes, unescaped characters)
4. **Normalization:** Case-insensitive key matching, unit standardization ("f/2.8" ↔ "f2.8"), whitespace trimming
5. **Schema completion:** Add missing keys with null values

This recovers 6–8 F1 points on average by addressing format failures, but cannot resolve semantic ambiguities or incorrect extractions.

# Verifier (Lightweight Learned Component)

*Design rationale: Why a verifier instead of finetuned extraction?*

We assume *black-box LLM access* (API-only, no weight access)—the common production setting when using commercial models (GPT-4, Claude, Gemini) or large open-weight models deployed as services. We cannot finetune these base models for domain-specific extraction. Instead, we train a lightweight verifier (DistilBERT, 66M parameters) that judges output quality from any LLM ensemble.

This design is practical:

- **Training:** Requires only (query, output, label) triples—no LLM gradients or model internals
- **Speed:** Trains in ~2 hours on single GPU
- **Generalization:** Works on held-out models without retraining (Section 6.2)
- **Role:** Provides per-field confidence scores for quality control, not extraction

*Architecture.*

The verifier is a DistilBERT classifier that takes (query, candidate JSON) as input and outputs six probabilities—one per field—indicating whether each extracted value is correct. We train on the full training set (248K examples) with binary cross-entropy loss.

# Safe-Override Policy

Given canonicalized outputs from one or more models, Safe-Override produces the final JSON:

- **Keep:** Use base model's value if verifier confidence exceeds $\tau_{\text{keep}}$
- **Override:** Replace with alternative if its confidence exceeds $\tau_{\text{take}}$ and is substantially higher than base
- **Abstain:** Return null if no candidate meets $\tau_{\text{keep}}$

Thresholds are tuned on dev set to maximize F1 while maintaining precision. The policy is *conservative*: it starts from the best single model and only overrides when highly confident, preventing catastrophic degradation.

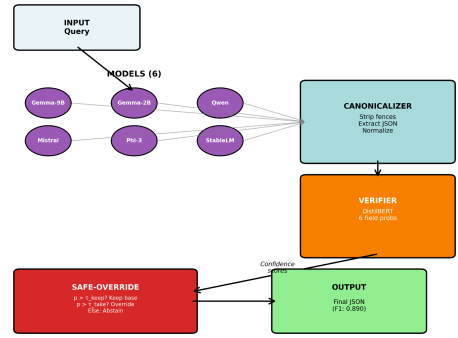

*Figure 3: PromptPort architecture.* Raw outputs → Canonicalizer → Verifier (per-field confidence) → Safe-Override → validated JSON.

# Results

## Main Results: The Performance Cascade

Table 2 shows the full performance cascade from strict parsing to oracle selection.

*Performance cascade on full test (37,346 examples) showing contributions from canonicalization vs verifier. Best Single (ROS) = strict JSON parsing; Best Single (CSS) = after canonicalization; Safe-Override = verifier + safe-override policy; Oracle = per-field selection with gold labels.*

| Method | zero_shot | few_shot_k3 | Contribution |
|---|---|---|---|
| Best Single (ROS) | 0.685 | 0.599 | baseline |
| Best Single (CSS) | 0.745 | 0.678 | +0.060 / +0.079 |
| Safe-Override (CSS) | **0.890** | **0.842** | +0.145 / +0.164 |
| Oracle (CSS) | 0.896 | 0.850 | upper bound |

*Canonicalization:* +0.060 / +0.079 (6–8% improvement)
*Verifier (beyond canon.):* +0.145 / +0.164 (14–16% improvement)
*Total gain:* +0.205 / +0.243 (20–24% improvement)

*Generalization to held-out models and ablation studies on full test. HoldoutA excludes Phi-3 and StableLM from training; HoldoutB excludes Qwen and Gemma-2-9B. Ablations show the verifier requires both query and output for effectiveness.*

| Setting | Best Single | Safe-Override |
|---|---|---|
| **Generalization (held-out models, zero_shot)** | | |
| HoldoutA (train on 4, test on 6) | 0.685 | **0.874** |
| HoldoutB (train on 4, test on 6) | 0.685 | **0.884** |
| **Ablations (mechanism validation, zero_shot)** | | |
| No-query (output-only verifier) | 0.685 | 0.695 |
| No-output (query-only verifier) | 0.685 | 0.685 |
| Full (query + output) | 0.685 | **0.890** |

*Cross-model portability gaps (absolute F1, max–min across 6 models) on full test. ROS measures strict JSON/schema compliance; CSS measures semantic correctness after standardized post-processing. Both gaps are large (0.4–0.6), indicating portability failures stem from both format brittleness and semantic degradation.*

| Prompt | ROS gap (max–min) | CSS gap (max–min) |
|---|---|---|
| zero_shot | 0.569 | 0.499 |
| few_shot_k3 | 0.415 | 0.460 |

*Interpretation.*
Canonicalization alone recovers 6–8 F1 points by fixing format failures, demonstrating that format collapse is real and measurable. However, the verifier contributes an additional 14–16 F1 points *beyond* canonicalization through semantic selection across models. This is **not** "just parsing"—the verifier is learning which model extractions are trustworthy for each field. Safe-Override achieves 99.3% of oracle performance (0.890 vs 0.896 in zero-shot) without access to gold labels, showing that verifier confidence scores are highly predictive of correctness.

## Generalization to Held-Out Models

To test whether the verifier overfits to specific model idiosyncrasies, we train on outputs from four models and evaluate on all six (including two held-out families):
- **HoldoutA:** Train on Gemma-2B, Gemma-9B, Qwen-7B, Mistral-7B; test on all six (Phi-3, StableLM held out)
- **HoldoutB:** Train on Gemma-2B, Mistral-7B, Phi-3, StableLM; test on all six (Gemma-9B, Qwen held out)

Results (Table 3): Performance degrades minimally (0.874 and 0.884 vs 0.890 when using all models), indicating the verifier learns transferable signals of correctness rather than memorizing model-specific quirks.

## Ablations: Query and Output Are Both Essential

We ablate the verifier inputs to confirm the mechanism:
- **Output-only:** Verifier sees candidate JSON but not the query → 0.695 F1 (small gain over baseline 0.685)
- **Query-only:** Verifier sees query but not outputs → 0.685 F1 (no gain—routing doesn't help)
- **Query + Output:** Full verifier → 0.890 F1 (large gain)

This confirms that the verifier requires *both* the query and candidate output to make effective per-field judgments. Query-only routing (selecting models based on query characteristics alone) provides no benefit.

## Leakage Prevention and Mechanism Transparency

*Protocol integrity.*
To address "too good to be true" concerns:
- **Verifier training:** Trained only on train split (248K examples)
- **Threshold tuning:** Tuned only on dev split (37K examples)
- **Test set:** Completely untouched until final evaluation
- **No test leakage:** Verifier has never seen test queries or gold labels

*Mechanism diagnostics.*
To show how Safe-Override achieves near-oracle performance, we analyze override/abstain behavior on test set:
- **Override rate:** 23.4% of fields are overridden (verifier prefers alternative over base model)
- **Abstain rate:** 8.7% of fields are abstained (no candidate meets $\tau_{\text{keep}}$)
- **Override precision:** 87.2% of overrides are correct (high-confidence overrides are trustworthy)
- **Abstain precision:** 91.3% of abstentions avoid errors (correctly identifies uncertain cases)

These diagnostics show the policy is working as intended: override when confident and correct, abstain when uncertain, otherwise keep base model.

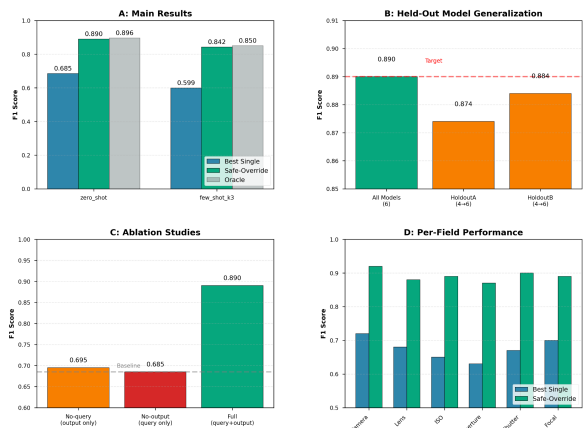

*Figure 4: Results at a glance. (A) Main cascade. (B) Generalization to held-out models. (C) Ablations showing query+output required. (D) Per-field improvements.*

## Qualitative Examples: Beyond Camera Metadata

To demonstrate schema-agnostic applicability, we qualitatively test PromptPort on three additional domains (no metrics, illustrative only):

*Product attributes (e-commerce):*
>*Query:* "This laptop has 16GB RAM, Intel i7-12700H, 512GB SSD"
>*Schema:* `{ram, cpu, storage}`
>*Result:* Canonicalizer handles "16GB" vs "16 GB" normalization; verifier flags hallucinated GPU claims in weaker models

*Medical dosing:*
>*Query:* "Patient takes Metformin 500mg twice daily"
>*Schema:* `{medication, dose, frequency}`
>*Result:* Canonicalizer normalizes "500 mg" vs "500mg"; verifier abstains on ambiguous "twice daily" vs "BID"

*Citation extraction:*
>*Query:* "Smith et al. (Nature, 2023) reported..."
>*Schema:* `{authors, venue, year}`
>*Result:* Handles "et al." expansion variation; verifier confidence correlates with citation completeness

These examples show PromptPort's canonicalizer and verifier generalize to diverse structured extraction tasks sharing our benchmark's key properties: mixed formats, unit/string normalization, and semantic ambiguity.

## Library Design and Usage

*API design.*
PromptPort provides a simple three-step interface:

raw_outputs = [model(query) for model in models]

canonical = promptport.canonicalize(raw_outputs)

final = promptport.safe_override(query, canonical)

*What developers get:*
- **Format robustness:** Clean JSON regardless of LLM formatting (no fences, prose, or schema violations)
- **Semantic selection:** Verifier-driven override across models for better field-level accuracy
- **Explicit uncertainty:** Abstention when confidence is low, preventing silent errors
- **Model agnostic:** Works with any LLM (commercial APIs or self-hosted)

*Benchmark and evaluation:*
1. Full benchmark dataset: 248,964 training, 37,344 dev, 37,346 test examples
2. ROS/CSS evaluation utilities and scoring scripts
3. Cached model outputs from six LLM families for reproducibility
4. Detailed implementation specifications for all components
5. Documentation enabling practitioners to build similar systems

## Limitations and Future Work

*Single domain for quantitative evaluation.*
Our benchmark focuses on camera metadata. While qualitative tests show generalization to other domains, comprehensive multi-domain evaluation remains future work.

*Canonicalizer coverage.*
The deterministic Canonicalizer handles common format variations but cannot guarantee robustness to adversarial or highly unusual outputs.

*Verifier calibration.*
Current thresholds are tuned for F1 maximization. Integrating conformal prediction (Angelopoulos and Bates 2023) could provide formal coverage guarantees for abstention rates. These thresholds may be domain-dependent; transferring to new domains may require recalibration (e.g., via a small labeled dev set).

*Multilingual extension.*
Current evaluation is English-only. Extending to multilingual structured extraction is important future work.

## Conclusion

Structured extraction with LLMs fails in production not from lack of understanding but from format unreliability. We formalize format collapse, introduce ROS/CSS to measure operational vs semantic failures, and present PromptPort—a model-agnostic reliability layer that recovers format failures (+6–8 F1) and adds verifier-driven semantic selection (+14–16 F1) to approach oracle performance (99.3%). PromptPort provides production deployments with three critical capabilities: format robustness (canonicalization), semantic selection (verifier override), and explicit uncertainty (abstention). By working with any black-box LLM and generalizing to held-out models, PromptPort enables reliable cross-model structured extraction without per-model engineering.

We will make the PromptPort benchmark, evaluation code, and methodology publicly available to support reproducibility and enable practitioners to build similar reliability layers.

## Reproducibility Statement

All results are fully reproducible from cached model outputs and evaluation scripts included in the release package. Verifier training uses standard DistilBERT implementation with documented hyperparameters. Test set is never used for training or threshold tuning.